\documentclass[
]{ceurart}

\usepackage{listings}
\usepackage{graphicx}
\usepackage{times}
\usepackage{latexsym}
\usepackage{multirow}
\usepackage{multicol}
\usepackage{tabularx}
\usepackage{graphicx}
\usepackage{booktabs}
\usepackage{mathtools}
\usepackage{enumitem}
\usepackage{booktabs}
\usepackage{multirow}
\usepackage{soul}

\usepackage{subcaption}
\usepackage{epstopdf}
\usepackage{xstring}

\usepackage{balance}
\usepackage{pbox}
\usepackage{multirow}
\usepackage{enumitem}
\usepackage{todonotes}
\usepackage[normalem]{ulem}
\usepackage{color}
\usepackage{tablefootnote}
\usepackage{multicol}
\usepackage{comment}

\begin{document}

\copyrightyear{2022}
\copyrightclause{Copyright for this paper by its authors.
  Use permitted under Creative Commons License Attribution 4.0
  International (CC BY 4.0).}

\conference{CLEF 2022: Conference and Labs of the Evaluation Forum, 
    September 5--8, 2022, Bologna, Italy}


\title{Z-Index at CheckThat! Lab 2022: Check-Worthiness Identification on Tweet Text}
% \title{Z-Index at CheckThat! 2022: BERT Models for Check-Worthiness Estimation on Tweet Text}

\author[1]{Prerona Tarannum}[%
orcid=0000-0002-3292-1870,
email=prerona15-14134@diu.edu.bd,
%url=https://yamadharma.github.io/,
]
\address[1]{Daffodil International University}

\author[1]{Md. Arid Hasan}[%
orcid=0000-0001-7916-614X,
email=arid.cse0325.c@diu.edu.bd,
%url=https://kmitd.github.io/ilaria/,
]
%\address[2]{Vrije Universiteit Amsterdam, De Boelelaan 1105, 1081 HV Amsterdam, The Netherlands}

\author[2]{Firoj Alam}[%
orcid=0000-0001-7172-1997,
email=fialam@hbku.edu.qa,
%url=http://conceptbase.sourceforge.net/mjf/,
]
\address[2]{Qatar Computing Research Institute}
\author[1]{Sheak Rashed Haider Noori}[%
orcid=0000-0001-6937-6039,
email=drnoori@daffodilvarsity.edu.bd,
%url=https://kmitd.github.io/ilaria/,
]

\begin{abstract}
The wide use of social media and digital technologies facilitates sharing %is responsible to spread 
various news and information about events and activities. Despite sharing positive information %, such enormous data, 
misleading and false information is also spreading on social media. % among the people. 
There have been efforts in identifying such misleading information both manually by human experts and automatic tools. Manual effort does not scale well due to the high volume of information, containing factual claims, are appearing online. Therefore, automatically identifying check-worthy claims can be very useful for human experts. %reduce the effort
% ?Estimating check-worthiness is one of the vital steps for identifying false and harmful information. 
In this study, we describe our participation in Subtask-1A: Check-worthiness of tweets (English, Dutch and Spanish) of CheckThat! lab at CLEF 2022. We performed standard preprocessing steps %on training data 
and applied different models %to English, Dutch, and Spanish tweet texts 
to identify whether a given text is worthy of fact-checking or not. We use the oversampling technique to balance the dataset and applied SVM and Random Forest (RF) with TF-IDF representations. We also used BERT multilingual (BERT-m) and XLM-RoBERTa-base pre-trained models for the experiments. We used BERT-m for the official submissions and our systems ranked as $3^{rd}$, $5^{th}$, and $12^{th}$ in Spanish, Dutch, and English, respectively. In further experiments, our evaluation shows that transformer models (BERT-m and XLM-RoBERTa-base) outperform the SVM and RF %classical models (SVM and RF) 
in Dutch and English languages where a different scenario is observed for Spanish.
%classical models outperform transformer models on Spanish language with respect to the positive class.
\end{abstract}


\maketitle

\section{Introduction}

% The rapid advancement of social media is one of the main streams of exchanging information among people.
Recently, social media became the main communication channel to exchanging information among people.
% As a result, relying on social media information 
As a result, it becomes the primary source of news \cite{perrin2020social}. In our daily activities, such information is helpful, however, a major part of them contains misleading content that is harmful to individuals, society, or organizations \cite{alam2020fighting,alam-etal-2021-fighting-covid}. 
The harmful or misleading content includes hate speech \citep{fortuna2018survey}, hostility \citep{brooke-2019-condescending,joksimovic-etal-2019-automated}, propagandistic news and memes \cite{da2020survey,EMNLP19DaSanMartino,dimitrov2021detecting,SemEval2021-6-Dimitrov}, harmful memes \cite{pramanick-etal-2021-momenta-multimodal}, abusive language \citep{mubarak2017abusive}, cyberbullying and cyber-aggression \citep{van-hee-etal-2015-detection,kumar2018benchmarking} and rumours \citep{bondielli2019survey}. 
% and the quality of this information is still in concern and 
The misleading or harmful aspects of such information raised the interest to identify and flag them to reduce their spread, further. There have been significant research efforts to automatically identify such content. Recent surveys on fake news \cite{zhou2020survey} disinformation \citep{alam2021survey}, rumours \citep{bondielli2019survey}, propaganda \citep{da2020survey}, multimodal memes  \citep{afridi2021multimodal}, hate speech \citep{fortuna2018survey}, 
cyberbullying \citep{7920246}, and offensive content \citep{husain2021survey} highlight the importance of the problem and relevant approaches to address them.

%questions about the spreading of disinformation.
% However, 
% Researchers has been working to find harmful and other disinformation which is shared through social media.
Most often information is disseminated with facts to make people believe it is true, which are typically found in political debates, and social and global agendas. Identifying whether such facts are true or false is an important step in fighting misleading information. There have been manual efforts by fact-checking organizations to identify the truthfulness of such factual statements. As such manual efforts do not scale well, therefore, it is important to automatically identify them. However, there is a reliability issue with the automated approach \cite{shaar2021assisting}. A trade-off is to support human fact-checkers using an automated approach, which includes different steps in the fact-checking pipeline~\cite{Survey:2021:AI:Fact-Checkers}.  
% As such content are abundant Finding disinformation requires to check the facts in the information and before checking the facts,  \cite{thorne2018automated}
The first step of the fact checking pipeline is to find content that is check-worthy. %iness estimation. 
The CheckThat! Lab (CTL) shared tasks is addressing this problem for the past several years. As an ongoing effort, this year CheckThat! Lab offered check worthiness subtask in six different languages such as Arabic, Bulgarian, Dutch, English, Spanish, and Turkish where data was collected from Twitter \cite{CheckThat:ECIR2022,clef-checkthat:2022:LNCS,clef-checkthat:2022:task1}. 
% As a results, we participated in a shared-task \cite{CheckThat:ECIR2022,clef-checkthat:2022:LNCS} organized by CheckThat! Lab (CTL) to identify the worthiness of fact-checking. 
% This year, CTL 2022 organized the Task 1 in six different languages including Arabic, Bulgarian, Dutch, English, Spanish (only for Subtask-1A), and Turkish where data was collected from twitter \cite{clef-checkthat:2022:task1}. 
We participated in check worthiness subtask and focused on Dutch, English, and Spanish. For the experiments, we used 
% In recent studies, 
different pretrained transformer based models, which have been widely used in several NLP tasks \cite{alam2020fighting, alam2021review}. %However, 
The difficulties arise when multilingual pretrained model is used in such tasks where facts and claims vary by country \cite{singh2022misinformation} and knowledge transferring across the language could spread the disinformation. We used multilingual transformer models (m-BERT and XLM-RoBERTa) for our experiments. 
% In this paper, we mainly focused on Subtask 1A in three languages (Dutch, English and Spanish) that we participated. For each language, we run four different algorithms which are SVM, RF, BERT-m, and XLM-RoBERTa. 
In addition to the transformer models, we also used SVM and RF with TF-IDF representations.

The rest of this paper is organized as follows. In section \ref{sec:related_work}, we provided related works that are relevant for this study. We then discuss the methodology
% describe the preprocessing steps, the classical models, and pre-trained BERT-based language models 
in Section \ref{sec:methodology}. Results of the experiments and detailed discussions are provided in Section \ref{sec:results}. Finally, we conclude our study in Section \ref{sec:conclusion}.

% \section{Background Study}
% \label{sec:background}

\section{Related Work}
\label{sec:related_work}

% \todo[inline]{
% Please check related work from the following papers: https://www.ijcai.org/proceedings/2021/0619.pdf \newline
% https://aclanthology.org/2021.findings-emnlp.56.pdf
% }
To deal with the factuality of statements there have been initiatives to manually check them and as result 
many fact-checking organizations have emerged, such as FactCheck.org\footnote{\url{http://www.factcheck.org/}},
Snopes\footnote{\url{http://www.snopes.com/fact-check/}},
PolitiFact\footnote{\url{http://www.politifact.com/}},
and FullFact\footnote{\url{http://fullfact.org/}}. In addition, there have also been some international initiatives such as the \textit{Credibility Coalition}\footnote{\url{https://credibilitycoalition.org/}} 
and \textit{Eufactcheck}\footnote{\url{https://eufactcheck.eu/}} \cite{stencel2019number}. 

One of the earlier efforts in this direction is the ClaimBuster system~\citep{Hassan:15}, which has been developed using the transcripts of 30 historical US election debates with a total of 28,029 transcribed sentences. The annotation includes \textit{non-factual}, \textit{unimportant factual}, and \textit{check-worthy factual} class labels and has been carried out by students, professors, and journalists. \citet{gencheva-EtAl:2017:RANLP} also focused on the 2016 US Presidential debates for which they obtained annotations from different fact-checking organizations. An extension of this work resulted in the development of ClaimRank, where the authors used more data and also included Arabic content \citet{NAACL2018:claimrank}. \citet{alam-etal-2021-fighting-covid} focused on COVID 19 topics in languages which are Arabic, Bulgarian, Dutch, and English, and achieved strong performances using pre-trained language models. The study also discussed the utility of single-task and multitask settings. 
The positive unlabelled learning technique for check-worthiness tasks has been introduced by \citet{wright2020claim} where authors experimented with this technique with the BERT model on different datasets and achieved the best results on two datasets out of three. The study of \citet{alhindi2021fact} introduced a multi-layer annotated news corpus and augmented discourse structure to understand the relation between fact-checking and argumentation. The first Turkish dataset for check-worthiness has been studied by \citet{kartal2020trclaim}, where BERT multilingual outperforms other models.

Some notable research outcomes came from shared tasks. For example, the CLEF CheckThat! labs' shared tasks \citep{clef2018checkthat:overall,clef-checkthat:2019,CheckThat:ECIR2019,shaar-etal-2021-findings} in the past few years featured challenges on automatic identification \citep{clef2018checkthat:task1,clef-checkthat-T1:2019} and verification \citep{clef2018checkthat:task2,clef-checkthat-T2:2019} of claims in political debates, and tweets \cite{clef-checkthat:2021:LNCS}.

%
% https://aclanthology.org/2020.findings-emnlp.43/
% https://aclanthology.org/2021.sigdial-1.40.pdf
% https://aclanthology.org/2020.conll-1.31/
% https://proceedings-of-deim.github.io/DEIM2022/papers/B34-4.pdf
% https://ieeexplore.ieee.org/stamp/stamp.jsp?arnumber=9677010&casa_token=mXIV7gxhAUkAAAAA:ZSgD8XALRgD8-7AAfZSvZB6-LpuusM215hpcLscoTlzxjhe8MS_lo9CcUNEdnv7SQYSHX4q7ODGz&tag=1

\section{Methodology}
\label{sec:methodology}

\subsection{Data}
\label{ssec:data}

\begin{figure}
\centering
\begin{subfigure}{.32\textwidth}
    \centering
    \includegraphics[width=.95\linewidth]{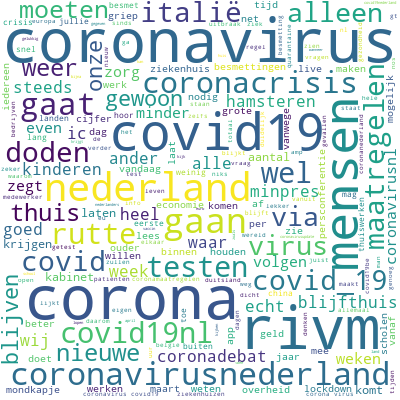}  
    \caption{Dutch}
    \label{sfig:dutch}
\end{subfigure}
\begin{subfigure}{.32\textwidth}
    \centering
    \includegraphics[width=.95\linewidth]{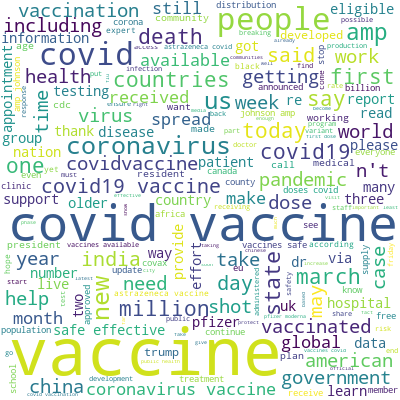}  
    \caption{English}
    \label{sfig:english}
\end{subfigure}
\begin{subfigure}{.32\textwidth}
    \centering
    \includegraphics[width=.95\linewidth]{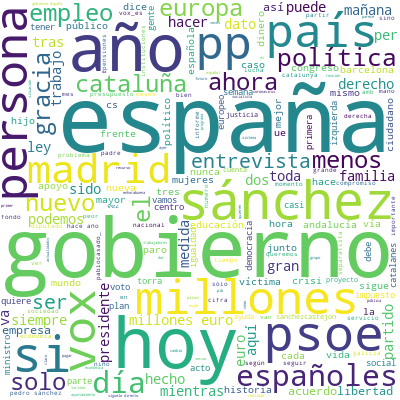}  
    \caption{Spanish}
    \label{sfig:spanish}
\end{subfigure}
\caption{Word-cloud representing top frequent words in different languages.
% (a) Word-cloud for Dutch language dataset (b) Word-cloud for English language dataset (c) Word-cloud for Spanish language dataset
}
\label{fig:wordcloud}
\end{figure}

The dataset we used in our study is obtained from CLEF CheckThat!2022 lab task1: \textit{Identifying Relevant Claims in Tweets} \cite{clef-checkthat:2022:task1}. The data is based on the COVID-19 topic for Dutch and English where Spanish is mixed of politics and COVID-19 topics, which is collected from Twitter. In table \ref{tab:dataset}, we present the distribution of the datasets that we used in this shared task to run our experiments. In Figure \ref{fig:wordcloud}, we present the word cloud for all three languages to understand the most common words present in the datasets. We first removed the stopwords from the data and then used the rest of the words to generate the most frequent words.

\begin{table}[]
\centering
\caption{Data splits and distributions of Subtask 1A: Check-worthiness of tweets}
%\vspace{-0.9em}
\label{tab:dataset}
%\scalebox{0.8}{
\begin{tabular}{@{}lrrrr@{}}
\toprule
\multicolumn{1}{c}{\textbf{Class label}} & \multicolumn{1}{c}{\textbf{Train}} & \multicolumn{1}{c}{\textbf{Dev}} & \multicolumn{1}{c}{\textbf{Test}} & \multicolumn{1}{c}{\textbf{Total}} \\ \midrule
\multicolumn{5}{c}{\textbf{Dutch}} \\ \midrule
No & 546 & 44 & 350 & 940 \\
Yes & 377 & 28 & 316 & 721 \\\cline{2-5}
\textbf{Total} & 923 & 72 & 666 & 1661 \\ \midrule
\multicolumn{5}{c}{\textbf{English}} \\ \midrule
No & 1675 & 151 & 110 & 1936 \\
Yes & 447 & 44 & 39 & 530 \\\cline{2-5}
\textbf{Total} & 2122 & 195 & 149 & 2466 \\ \midrule
\multicolumn{5}{c}{\textbf{Spanish}} \\ \midrule
No & 3087 & 2195 & 4296 & 9578 \\
Yes & 1903 & 305 & 704 & 2912 \\\cline{2-5}
\textbf{Total} & 4990 & 2500 & 5000 & 12490 \\
 \bottomrule
\end{tabular}
%}
%\vspace{-0.3cm}
\end{table}

%\fa{please discuss whether you used all words? or filtered them?}
% Most common words are COVID-19 related for Dutch and English languages we can identify from the figures. 

\subsection{Preprocessing}
\label{ssec:preprocess}

% The data we used in our study 
The CTL subtask-1A datasets are collected from Twitter. As a result, the data contains many symbols, URLs, and invisible characters. We performed several preprocessing steps to clean the noisy data. First, we perform URLs and unnecessary character removal steps by following the approach discussed in \cite{alam2020standardizing}. Then, we removed the stopwords from the data. Finally, we removed hashtag signs and usernames.

% \subsection{Model and Architecture}
\subsection{Models}
\label{ssec:model}
We used both deep learning and traditional models to run classification experiments. 
As deep learning algorithms, we used two transformer based models, BERT~\cite{devlin2018bert} and XLM-RoBERTa \cite{conneau2019unsupervised}. Several factors were considered while choosing the algorithms. Among the transformer based models, BERT and XLM-RoBERTa are larger in parameter size.\footnote{110 million parameters in \textit{BERT multilingual} and 125 million in \textit{XLM-RoBERTa base}} The number of parameters and network size is responsible for computation time and performance of the learning. For these two models, we used the multilingual version of the models.
For the later case, we used the two most popular algorithms such as \textit{(i)} Random Forest (RF) ~\cite{liaw2002classification}, and \textit{(ii)} Support Vector Machines (SVM) ~\cite{platt1998sequential}.
% as part of classical algorithms.

\subsection{Experiments}
\label{ssec:experiments}
\paragraph{Transformers models} We use the Transformer Toolkit~\cite{Wolf2019HuggingFacesTS} for transformer-based models. We used learning rate of $1e-5$ to fine-tune each model \cite{devlin2018bert}. Model specific tokenizer is available with Transformer Toolkit that we used in our study. For transformer based model, we run $4$, $2$, and $8$ epochs for BERT-m model for Dutch, English, and Spanish languages, and $4$, $4$, and $8$ epochs for Dutch, English, and Spanish languages for XLM-RoBERTa-base model.
% Random seed $42$ is used for all transformer based models.
\begin{table}[]
\caption{Hyper-parameters for traditional models to reproduce the results.
}
%\vspace{-0.9em}
\label{tab:params_classical}
\begin{tabular}{lrrrrrr}
\toprule
\multicolumn{1}{c}{\multirow{2}{*}{\textbf{Parameters}}} &
  \multicolumn{2}{c}{\textbf{Dutch}} &
  \multicolumn{2}{c}{\textbf{English}} &
  \multicolumn{2}{c}{\textbf{Spanish}} \\ \cmidrule{2-7} 
\multicolumn{1}{c}{} &
  \multicolumn{1}{c}{\textbf{SVM}} &
  \multicolumn{1}{c}{\textbf{RF}} &
  \multicolumn{1}{c}{\textbf{SVM}} &
  \multicolumn{1}{c}{\textbf{RF}} &
  \multicolumn{1}{c}{\textbf{SVM}} &
  \multicolumn{1}{c}{\textbf{RF}} \\ \midrule
Number of Feature & 1850 & 1500 & 1750 & 2800 & 3200 & 1700 \\
N-gram            & 3    & 3    & 4    & 3    & 4    & 3    \\
Random Seed       & 2814 & 2814 & 2814 & 2814 & 2814 & 2814 \\\bottomrule
\end{tabular}
\end{table}

\paragraph{Traditional Algorithms} To train the classifiers using the above-mentioned traditional models, we first transformed the preprocessed data into tf-idf vectors with weighted $n$-gram (unigram, bigram and trigram) to use contextual information. The class distribution of provided dataset for English and Spanish is not well balanced. Therefore, to balance the class distribution, we applied oversampling techniques \cite{chawla2002smote} for all three languages.
%Fianlly, we applied grid search for parameter optimization for both SVM and RF.
% \rev{
% 
We merged the train and dev-test set to train the model. 
We applied the upsampling technique to the combined dataset with a ratio of 1.0 with respect to the negative class. In Table \ref{tab:params_classical}, we report the hyper-parameters with the values to reproduce our results.
% }

%\fa{
%Please write more detail about experiments, like whether all data used to train the model? like train, dev and dev-test? if you have any results on dev-test you can consider reporting the results too. 
%For transformer based model, how many epoch you trained on ? random seed? just one run? 
%For the upsampling, what propotion you upsampled? 
%}

\section{Results and Discussion}
\label{sec:results}

% \todo[inline]{FA: detail results are good, but please report official results too whatever is the ranking...
% Arid: Reported Sir.
% }

In Table \ref{tab:official_results}, we report the official results and ranking evaluated by the lab organizers. The official evaluation metric for subtask 1A is F1 measure with respect to the positive class.

In Table \ref{tab:result}, we report the detailed classification results for each language. After releasing the gold set once the submission period ends, we re-run all the experiments and reported the detailed results. From the table, we can conclude that among the traditional models the performance of SVM is much better than RF except for Spanish data where RF is 0.25\% higher. 
% \rev{
The upsampling technique for traditional models improves from 0.10\% to 1.10\% %F1 score 0.10\%--1.10\% 
on different languages with respect to the positive class.
% }
We know from the literature, transformer based models are well-known for their performances and capabilities. Although XLM-Roberta base and BERT-m models provide the best results for Dutch and English languages with respect to positive class, where the traditional model outperforms the transformer models on Spanish language by a large margin. 

\begin{table}[]
\centering
\caption{Official results on the test set and overall ranking of Subtask 1A: Check-worthiness of tweets}
%\vspace{-0.9em}
\label{tab:official_results}
%\scalebox{0.8}{
\begin{tabular}{@{}llrr@{}}
\toprule
\multicolumn{1}{c}{\textbf{Language}} & \multicolumn{1}{c}{\textbf{Model}} & \multicolumn{1}{c}{\textbf{F1 (postive class)}} & \multicolumn{1}{c}{\textbf{Rank}} \\ \midrule
Dutch & BERT-m & 0.497 & $5^{th}$ \\ \midrule
English & BERT-m & 0.478 & $12^{th}$ \\ \midrule %\cline{2-5}
Spanish & BERT-m & 0.303 & $3^{rd}$ \\
 \bottomrule
\end{tabular}
%}
%\vspace{-0.3cm}
\end{table}

\begin{table}[]
\centering
\caption{Detail results on the test set of Subtask 1A: Check-worthiness of tweets. \textbf{Bold} indicates positive class F1 score. \ul{\textit{Underline}} indicates best F1 score for each language.
%\rev{please use bold form for }
}
%\vspace{-0.9em}
\label{tab:result}
%\scalebox{0.8}{
\begin{tabular}{@{}llrrrr@{}}
\toprule
\multicolumn{1}{c}{\textbf{Class label}} & \multicolumn{1}{c}{\textbf{Model}} & \multicolumn{1}{c}{\textbf{Accuracy}} &  \multicolumn{1}{c}{\textbf{Precision}} & \multicolumn{1}{c}{\textbf{Recall}} & \multicolumn{1}{c}{\textbf{F1 Score}}\\ \midrule
\multicolumn{6}{c}{\textbf{Dutch}} \\ \midrule
%SVM
\multicolumn{1}{c}{No} & \multirow{2}{*}{SVM} & \multirow{2}{*}{59.01} &  \multicolumn{1}{c}{60.85} & \multicolumn{1}{c}{61.71} & \multicolumn{1}{c}{61.28}\\ \cline{4-6}
\multicolumn{1}{c}{Yes} & & & \multicolumn{1}{c}{56.91} & \multicolumn{1}{c}{56.01} & \multicolumn{1}{c}{\textbf{56.46}}\\ 
\midrule
%RF
\multicolumn{1}{c}{No} & \multirow{2}{*}{RF} & \multirow{2}{*}{57.96} &  \multicolumn{1}{c}{57.85} & \multicolumn{1}{c}{73.71} & \multicolumn{1}{c}{64.82}\\ \cline{4-6}
\multicolumn{1}{c}{Yes} & & & \multicolumn{1}{c}{58.18} & \multicolumn{1}{c}{40.51} & \multicolumn{1}{c}{\textbf{47.76}}\\ 
\midrule
%BERT-m

\multicolumn{1}{c}{No} & \multirow{2}{*}{BERT-m} & \multirow{2}{*}{60.06} &  \multicolumn{1}{c}{60.82} & \multicolumn{1}{c}{67.43} & \multicolumn{1}{c}{63.96}\\ \cline{4-6}
\multicolumn{1}{c}{Yes} & & & \multicolumn{1}{c}{58.99} & \multicolumn{1}{c}{51.90} & \multicolumn{1}{c}{\textbf{55.22}}\\ 
\midrule
%RoBERTa-base

\multicolumn{1}{c}{No} & \multirow{2}{*}{XLM-RoBERTa base} & \multirow{2}{*}{56.76} &  \multicolumn{1}{c}{60.00} & \multicolumn{1}{c}{53.14} & \multicolumn{1}{c}{56.36}\\ \cline{4-6}
\multicolumn{1}{c}{Yes} & & & \multicolumn{1}{c}{53.93} & \multicolumn{1}{c}{60.76} & \multicolumn{1}{c}{\textbf{\ul{57.14}}}\\ 
\midrule

\multicolumn{6}{c}{\textbf{English}} \\ \midrule

\multicolumn{1}{c}{No} & \multirow{2}{*}{SVM} & \multirow{2}{*}{69.80} &  \multicolumn{1}{c}{85.71} & \multicolumn{1}{c}{70.91} & \multicolumn{1}{c}{77.61}\\ \cline{4-6}
\multicolumn{1}{c}{Yes} & & & \multicolumn{1}{c}{44.83} & \multicolumn{1}{c}{66.67} & \multicolumn{1}{c}{\textbf{53.61}}\\ 
\midrule

\multicolumn{1}{c}{No} & \multirow{2}{*}{RF} & \multirow{2}{*}{75.17} &  \multicolumn{1}{c}{76.64} & \multicolumn{1}{c}{95.45} & \multicolumn{1}{c}{85.02}\\ \cline{4-6}
\multicolumn{1}{c}{Yes} & & & \multicolumn{1}{c}{58.33} & \multicolumn{1}{c}{17.95} & \multicolumn{1}{c}{\textbf{27.45}}\\ 
\midrule

\multicolumn{1}{c}{No} & \multirow{2}{*}{BERT-m} & \multirow{2}{*}{63.09} &  \multicolumn{1}{c}{89.86} & \multicolumn{1}{c}{56.36} & \multicolumn{1}{c}{69.27}\\ \cline{4-6}
\multicolumn{1}{c}{Yes} & & & \multicolumn{1}{c}{40.00} & \multicolumn{1}{c}{82.05} & \multicolumn{1}{c}{\textbf{\ul{53.78}}}\\ 
\midrule

\multicolumn{1}{c}{No} & \multirow{2}{*}{XLM-RoBERTa base} & \multirow{2}{*}{51.01} &  \multicolumn{1}{c}{91.11} & \multicolumn{1}{c}{37.27} & \multicolumn{1}{c}{52.90}\\ \cline{4-6}
\multicolumn{1}{c}{Yes} & & & \multicolumn{1}{c}{33.65} & \multicolumn{1}{c}{89.74} & \multicolumn{1}{c}{\textbf{48.95}}\\ 
\midrule

\multicolumn{6}{c}{\textbf{Spanish}} \\ \midrule

\multicolumn{1}{c}{No} & \multirow{2}{*}{SVM} & \multirow{2}{*}{84.76} &  \multicolumn{1}{c}{92.89} & \multicolumn{1}{c}{89.08} & \multicolumn{1}{c}{90.95}\\ \cline{4-6}
\multicolumn{1}{c}{Yes} & & & \multicolumn{1}{c}{46.70} & \multicolumn{1}{c}{58.38} & \multicolumn{1}{c}{\textbf{51.89}}\\ 
\midrule

\multicolumn{1}{c}{No} & \multirow{2}{*}{RF} & \multirow{2}{*}{88.62} &  \multicolumn{1}{c}{91.27} & \multicolumn{1}{c}{95.93} & \multicolumn{1}{c}{93.54}\\ \cline{4-6}
\multicolumn{1}{c}{Yes} & & & \multicolumn{1}{c}{63.92} & \multicolumn{1}{c}{44.03} & \multicolumn{1}{c}{\textbf{\ul{52.14}}}\\ 
\midrule

\multicolumn{1}{c}{No} & \multirow{2}{*}{BERT-m} & \multirow{2}{*}{68.30} &  \multicolumn{1}{c}{91.75} & \multicolumn{1}{c}{69.34} & \multicolumn{1}{c}{78.99}\\ \cline{4-6}
\multicolumn{1}{c}{Yes} & & & \multicolumn{1}{c}{24.87} & \multicolumn{1}{c}{61.93} & \multicolumn{1}{c}{\textbf{35.49}}\\ 
\midrule
\multicolumn{1}{c}{No} & \multirow{2}{*}{XLM-RoBERTa base} & \multirow{2}{*}{70.64} &  \multicolumn{1}{c}{90.33} & \multicolumn{1}{c}{73.72} & \multicolumn{1}{c}{81.18}\\ \cline{4-6}
\multicolumn{1}{c}{Yes} & & & \multicolumn{1}{c}{24.43} & \multicolumn{1}{c}{51.85} & \multicolumn{1}{c}{\textbf{33.21}}\\ 
 \bottomrule
\end{tabular}
%}
%\vspace{-0.3cm}
\end{table}

%We receive the lowest F1 score with respect to positive class for each language data using XLM-RoBERTa large. The table shows that XLM-RoBERTa large couldn't classify any positive class data for Dutch and English.

%\rev{Please discuss any findings due to the upsampling.... how much did it improved the performance?}

\section{Conclusion}
\label{sec:conclusion}

In this study, we have run comparative experiments using different check-worthiness claim datasets consisting of Dutch, English, and Spanish languages, which are provided by CLEF CheckThat! lab 2022 organizers as a part of shared tasks. We cleaned the data to run the classification experiments. We investigated different machine learning algorithms including traditional (i.e., SVM) and deep learning models (i.e., BERT multilingual). Despite the cost of increased resource and time complexity, transformer based models did not perform well for Spanish language, however, outperformed the Dutch and English languages. Our study reveals that the transformer based models outperforms the traditional machine learning approach for Dutch and English language tasks.
% Larger models in terms of number of parameters performs significantly lower, i.e., \textit{XLM-RoBERTa large}.

\section{Acknowledgments}
We would like to thank the organizers and other participants in the challenge. We are thankful to DIU NLP and ML Research Lab for the workplace support. Finally, thanks to all the anonymous reviewers for their suggestions.

Part of this work is made within the Tanbih mega-project,\footnote{\url{http://tanbih.qcri.org}} developed at the Qatar Computing Research Institute, HBKU, which aims to limit the impact of ``fake news'', propaganda, and media bias by making users aware of what they are reading, thus promoting media literacy and critical thinking.

% \begin{acknowledgments}
%   Thanks to the developers of ACM consolidated LaTeX styles
%   \url{https://github.com/borisveytsman/acmart} and to the developers
%   of Elsevier updated \LaTeX{} templates
%   \url{https://www.ctan.org/tex-archive/macros/latex/contrib/els-cas-templates}.  
% \end{acknowledgments}

%%
%% Define the bibliography file to be used
\bibliography{main.bib}

\end{document}